\documentclass[letterpaper, 10 pt, conference]{ieeeconf}  

\IEEEoverridecommandlockouts                              

\usepackage{xcolor}
\usepackage[linesnumbered,ruled,vlined]{algorithm2e}
\usepackage{amssymb}
\usepackage{amsmath}
\usepackage{graphicx}
\graphicspath{ {./images/} }

\SetCommentSty{mycommfont}
\usepackage[linesnumbered,ruled,vlined]{algorithm2e}
\usepackage{algpseudocode}
\usepackage{caption}
\usepackage{subcaption}
\usepackage[utf8]{inputenc}
\usepackage{hyperref}
\usepackage{float}

\usepackage{cite}

\SetKwInput{KwInput}{Input}
\SetKwInput{KwOutput}{Output}

\title{\LARGE \bf
Motion Primitives Based Kinodynamic RRT for Autonomous Vehicle Navigation in Complex Environments
}

\author{Shubham Kedia$^{1}$ and Sambhu H. Karumanchi$^{2}$
\thanks{$^{1}$Shubham Kedia is with the Department of Mechanical Science and Engineering, University of Illinois at Urbana-Champaign,
        IL, USA 
        {\tt\small skedia4@illinois.edu}}%
\thanks{$^{2}$Sambhu H Karumanchi is with the Department of Aerospace Engineering, University of Illinois at Urbana-Champaign,
        IL, USA 
        {\tt\small shk9@illinois.edu}}%
}

\begin{document}

\maketitle

\begin{abstract}
 In this work, we have implemented a SLAM-assisted navigation module for a real autonomous vehicle with unknown dynamics. The navigation objective is to reach a desired goal configuration  along a collision-free trajectory while adhering to the dynamics of the system. Specifically, we use LiDAR-based Hector SLAM for building the map of the environment, detecting obstacles, and for tracking vehicle's conformance to the trajectory as it passes through various states. For motion planning, we use rapidly exploring random trees (RRTs) on a set of generated motion primitives to search for dynamically feasible trajectory sequences and collision-free path to the goal. We demonstrate complex maneuvers such as parallel parking, perpendicular parking, and reversing motion by the real vehicle in a constrained environment using the presented approach. The demo videos are available \href{https://drive.google.com/drive/folders/1NzYhRjUM0KSE0MDjicWkTqcjokVSojSz?usp=sharing}{HERE}.
\end{abstract}

\section{INTRODUCTION}
Autonomous vehicles need reliable perception and robust motion planning to perform complex maneuvers while avoiding obstacles in their operating environment. Trajectories generated by basic path planners that model only kinematics are often non-executable in the real world because of the limits on actuator forces and torques. Kinodynamic planning \cite{LaVelle2001}, by design, integrates system dynamics constraints of the form $\dot{x} = f(x,u)$ in its modeling to guarantee collision-free trajectories, where $x \in \mathbb{R}^n$ represents the vehicle state, and $ u \in \mathbb{R}^m$ the control action. Kinodynamic path planning has been actively researched for decades for robot navigation. \cite{donald1993kinodynamic, hsu2002randomized, lau2009kinodynamic, ardiyanto2012real}. 

In real world, it is quite challenging to model the system dynamics as an explicit mathematical function $\dot{x} = f(x,u)$. Especially, for autonomous vehicles the system identification is non-trivial due to the non-linearities in actuators and vehicle systems. \cite{sheetzsystem, ducaju2020application, karkoub2021system} explored high fidelity autonomous vehicle dynamics identification. But their applicability for real-time navigation tasks is limited due to high model complexity.

Also, planning has to deal with the curse of dimensionality because it entails search in a high dimension configuration space. The need for search over a high dimensional space popularized the randomized sampling based Rapidly exploring Randomized Trees (RRTs) \cite{lavalle2006planning} algorithm which, starting with an initial vertex, incrementally constructs the tree by repeatedly sampling a point in the configuration space, finding its nearest neighbor vertex in the existing tree, and applying an allowable control that pulls the vertex towards the random point. The latter feature biases the growth of the tree towards unexplored space and returns a solution as soon as the tree reaches the goal, a feature that makes the algorithms suitable for a fast online implementation.

\begin{figure}[!ht]
\centering
  \includegraphics[width=8cm]{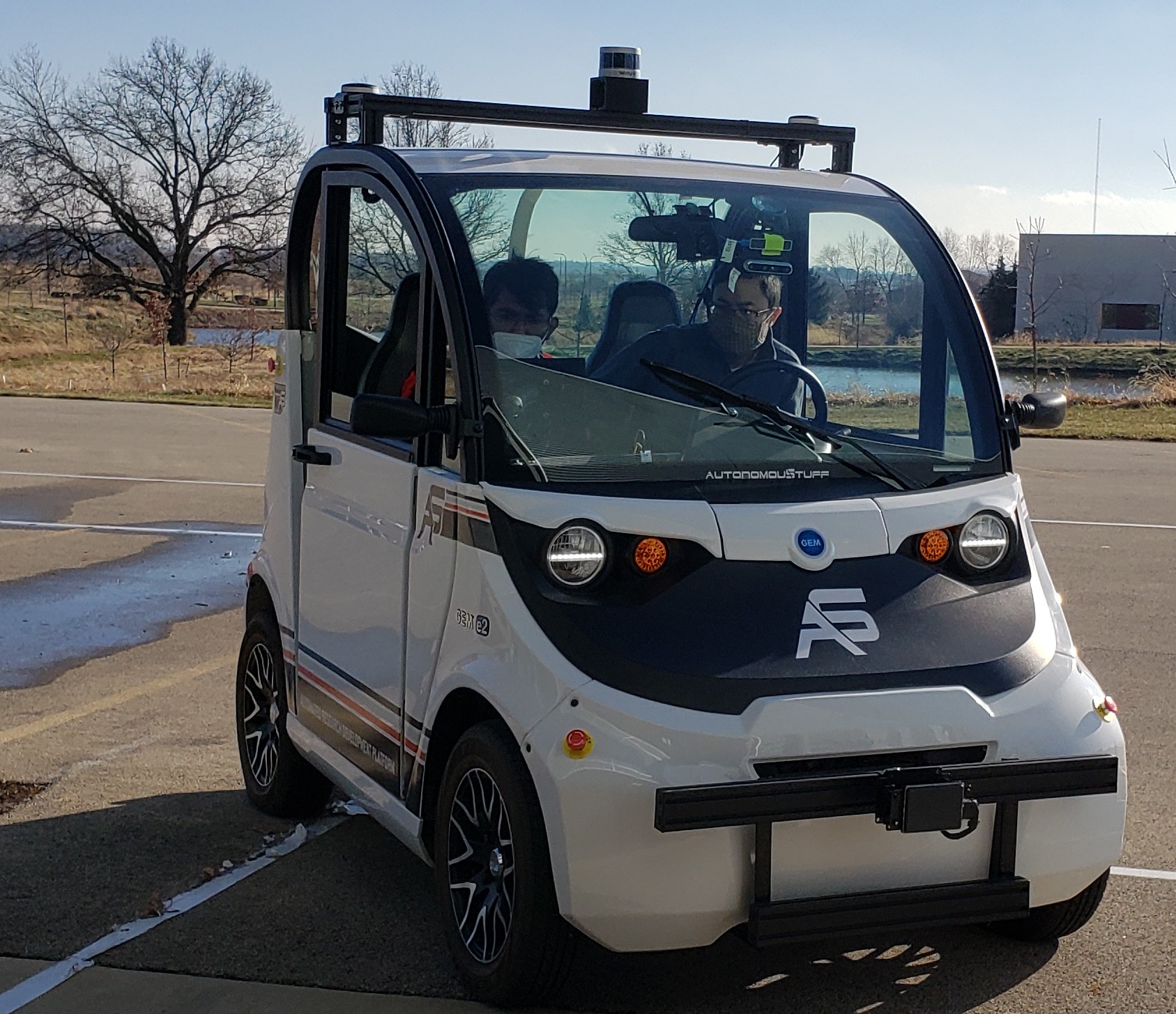}
  \caption{Polaris GEM e2 autonomous vehicle }
  \label{fig:vehicle}
\end{figure}

In this approach, a set of discrete control sequences called motion primitives are used to extend the tree by appending the node that is closest to the new sampled configuration at every step. We implement motion primitives based RRT on a real autonomous vehicle and demonstrate complex maneuvers like parallel parking, perpendicular parking, and reversing in a constrained environment. The vehicle is Polaris GEM e2 (Figure \ref{fig:vehicle}), a small, certified autonomous vehicle housed at the Center for Autonomy at the University of Illinois, Urbana-Champaign. We generate a fixed set of motion primitives (described in detail in Section 3) by performing various different driving maneuvers on the autonomous car to select a few simple, plausible maneuvers as our motion primitives. The motion primitives based approaches have the advantage that they can be so designed to directly encode various basic motion behaviors depending on the application and generate feasible trajectories with the help of sampling algorithms like RRT. Refer to \cite{Henderson2021} for an interesting application to the socially-aware robot navigation where socially-guided primitives enact safe interaction between humans and agents. Thus, motion primitives based RRT offers a flexible framework for autonomous vehicle motion planning in diverse environments. In a vast majority of the works, the use of RRTs in on-line planning systems for robotic vehicles has been restricted to simulation, or to kinematics. As for real-world implementations, MIT DARPA Urban Challenge vehicle \cite{Kuwata2008} used on-line RRT to deal with complex, unstable dynamics and drift. In the same vein, our work reports a more flexible {\em motion primitives based RRT} on a real-world autonomous car albeit in a laboratory environment.  

The ease of autonomous vehicle navigation is tightly coupled to reliable and accurate perception of its environment. Our autonomous car uses LiDAR scans to perceive the map of the environment and localizes itself within the map using SLAM (Simultaneous Localization and Mapping). We specifically use Hector SLAM \cite {Kohlbrecher} in our implementation because of its ability to match with the high update rate of LiDAR systems and provide accurate 2D pose estimates. While Hector SLAM system does not aim at explicit loop closing, it is known to generate sufficiently accurate mapping in many settings \cite{Shahi2020}. In Section III, we describe the overall methodology in detail.

\section{RELATED WORKS}
RRT \cite{LaVelle1998} is a pioneering contribution in robot motion planning which, along with its many important extensions, found pervasive use in many robotic applications. Since it will be hard to review all chronological contributions, we will review some recent works pertaining only to autonomous vehicles and urge the reader to refer to \cite{Noreen2016} for a comprehensive survey on RRT related works, and  \cite{Gonzales2016} and \cite{Katrakazas2015} for a review of motion planning techniques for autonomous vehicles, in general. \cite{Ma2014} proposed a fast RRT algorithm for autonomous road vehicles that uses off-line templates of the traffic scenes to assist in tree search. \cite{Shin2016} invokes desired orientation during the phase of tree expansion to address navigation challenges in cluttered places like parking lots. \cite{Garrote2014} focuses on autonomous navigation and collision detection using a RRT-based dynamic path planning and a path-following controller, and provides results in a simulated environment.\cite{Niclas2015} proposes a sampling-based Closed-Loop Rapidly exploring Random Tree (CL-RRT) for autonomous heavy duty vehicles which often have second order differential constraints.

Authors Pivtoraiko et. al. \cite{pivtoraiko2009differentially} described motion primitives based state lattice planner. The primitives are designed via sampling in state spaces and are used to perform incremental search using the $D^*$ Lite \cite{koenig2002d} algorithm. \cite{butzke2014state} also explores search based planning using state lattice and controller-based motion primitives. However, these approaches are limited to simulation and the proposed path planning cannot be implemented on a real autonomous vehicle with unknown dynamics. \cite{grymin2014hierarchical} and \cite{likhachev2009planning} are closest to our approach. \cite{grymin2014hierarchical} discusses about path planning by concatenation of pre-specified motion primitives and also studied graph-based search techniques, where the graph does not represent a state lattice but rather exhibits a tree structure. \cite{likhachev2009planning} presents  incremental search on a multi-resolution, dynamically feasible lattice state space and includes experimental validation on real autonomous vehicle but it employs an accurate vehicle model.

In relation to the above works, our primary contributions are as follows:
\begin{itemize}
    \item The primary focus of our work is to depart from simulated environment and demonstrate actual implementation of RRT-based motion planning and trajectory tracking on a real autonomous vehicle.
    \item In contrast to the other RRT-based works, our approach is designed to operate under unknown dynamics of the vehicle. Moreover,  our study is motion primitives-centric that allows for encoding of application-specific constraints such as actuator limits, for example, while generating feasible trajectories.
    \item Our real-world experiments demonstrate the feasibility of performing complex maneuvers like parallel parking, perpendicular parking, and reversing the vehicle using a single set of motion primitives within the RRT algorithm.
\end{itemize}

\section{METHODOLOGY}
\subsection{Mapping and Localization}
To find a collision-free trajectory from the start position to the goal position, the autonomous car must create a map of the environment and dynamically register its position as it moves within the environment. Our autonomous car uses 2D LiDAR scans to perceive the environment and localizes itself using the Hector SLAM algorithm.  
 For completeness, we present here a few important computations involved in Hector SLAM. Primarily, Hector SLAM uses an occupancy grid as the map of the environment where each cell of the grid is in one of the three states-{\em occupied, free, or unexplored}. A LiDAR measurement is assigned $1,-1$ when the LiDAR hits occlusion, or passes through the free space respectively. The cells unexplored in the map by LiDAR are set to $0$. 

The problem of scan matching (matching the current scan either with a previous scan or with an existing map)  is to find the rigid transformation $\xi = (p_x, p_y, \theta)$ that minimizes
\begin{equation}
\xi^* = \mbox{argmax}_ \xi \sum_{i=1}^{n} [1-M(S_i(\xi)]^2
\end{equation}
where $S_i(\xi)$ are the world coordinates of the endpoints of scan $i$, $ s_i = \begin{pmatrix} 
                                                                               s_{i,x}\\
                                                                                s_{i,y}
                                                                                \end{pmatrix} $ given by
\begin{equation}
   S_i(\xi) =  \begin{pmatrix}
                   cos(\psi) & sin(-\psi) \\
                   sin(\psi) & cos(\psi)
                   \end{pmatrix}  \begin{pmatrix}
                                    s_{i,x} \\
                                    s_{i,y}
                                \end{pmatrix} + \begin{pmatrix}
                                                  p_x \\
                                                  p_y
                                                \end{pmatrix} \label{Sis}
\end{equation}

In the above, $M(S_i(\xi))$ is the map value corresponding to the point $S_i(\xi))$. In terms of differential movement, the above minimization problem can be transformed into finding a change in pose $\Delta \xi$ such that 
\begin{equation}
 \sum_{i=1}^{n} [1-M(S_i(\xi + \Delta \xi)]^2  \rightarrow 0  \label{diffOpt}
\end{equation}
Applying the first order Taylor's approximation, one can show that (\ref{diffOpt}) is optimized at
\begin{equation}
\Delta \xi  = H^{-1} \sum_{i=1}^{n}\left[ \nabla M(S_i(\xi))\frac{\partial{S_i(\xi)}}{\partial \xi} \right] ^T\left[ 1-M(S_i(\xi)) \right] \label{opteq}
\end{equation}
where $H = \left[ \nabla M(S_i(\xi))\frac{\partial{S_i(\xi)}}{\partial \xi} \right] ^T \left[ \nabla M(S_i(\xi))\frac{\partial{S_i(\xi)}}{\partial \xi} \right] $.

Using (\ref{Sis}), $\frac{\partial{S_i(\xi)}}{\partial \xi}$ can be written as
\begin{equation}
    \frac{\partial{S_i(\xi)}}{\partial \xi}  = \begin{pmatrix}
                                                1 & 0 & -sin(\psi)s_{i,x} & -cos(\psi)s_{i,y} \\
                                                0 & 1 & cos(\psi)s_{i,x} & -sin(\psi)s_{i,y}
                                                \end{pmatrix}
\end{equation}
and the changes in map values, or equivalently the gradient $\nabla M(S_i(\xi))$ is computed by applying linear interpolation between the cell values of the occupancy grid at the current estimate $\xi$. Thus, (\ref{opteq}) can be used to get the optimal pose change estimate. The procedure is repeated to get the full map of the environment surrounding the robot which will be used below to identify the right trajectory for obstacle avoidance. 

\subsection{Motion primitives generation}
\begin{figure}
    \centering
    \includegraphics[width = 9cm]{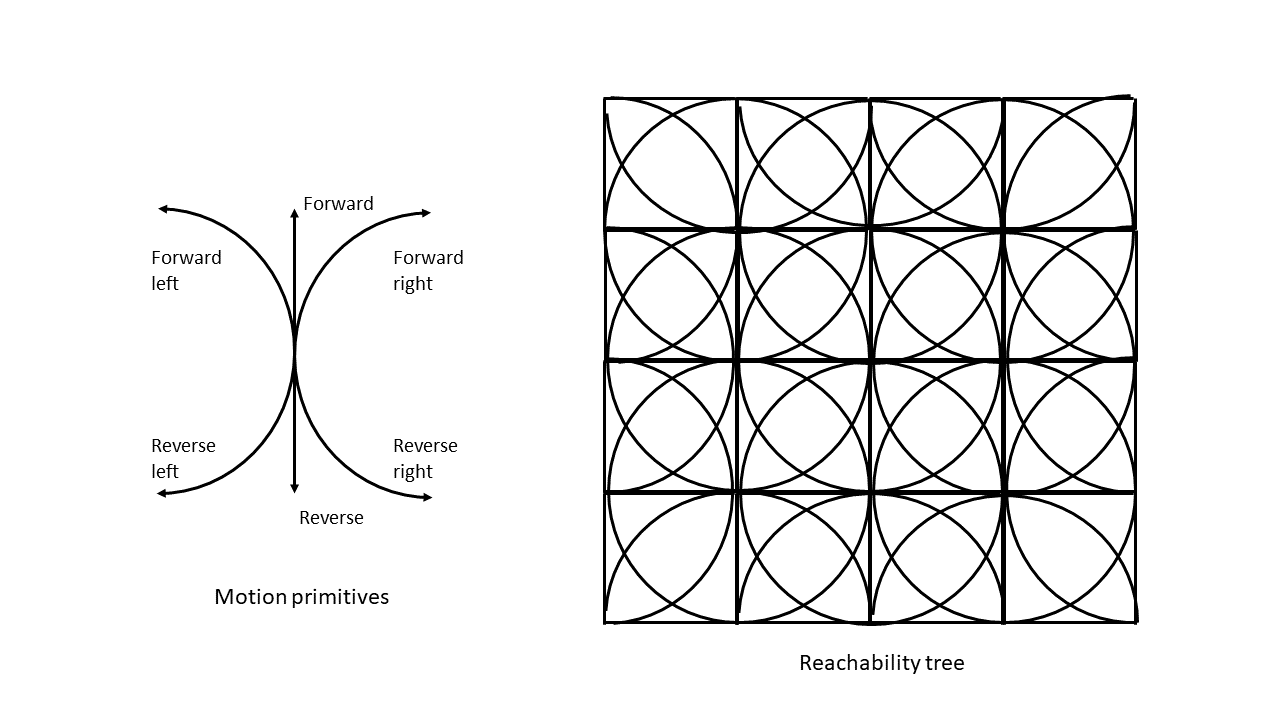}
    \caption{Motion primitives and reachability tree}
    \label{MotionPrimitives}
\end{figure}

A trajectory is a series of adjacent states in a graph. However, not all geometrically adjacent points are negotiable by the robot because of the non-holonomic constraints such as impossibility of side-wise movement by the car. Hence, to facilitate simplified feasible control, we assume that that the motion of the autonomous car is executed in terms of certain {\it motion primitives}, which are sequence of control and state pairs generated by recording the control inputs during manual driving and vehicle state tracking with the SLAM.
We allow the autonomous car to use the six motion primitives depicted in the figure (\ref{MotionPrimitives}) for its maneuvers. The design for motion primitives is based on Reeds-Sheeps curves \cite{reeds1990optimal} and for simplicity only six primitives are used but the technique allows extrapolation to a higher number of primitives for a more optimal trajectory. Assuming full steering rotation as allowed by the design, the primitives considered here are  {\it forward straight, reverse straight, forward right curve, forward left curve, reverse right curve, reverse left curve}. The motion primitives and the reachability tree possible through these primitives are shown in figure (\ref{MotionPrimitives}).

\subsection{Kinodynamic RRT planner}

Let $ Z \in\mathbb{R}^4 $ denote the configuration space and $U \in\mathbb{R}^4$ denote the control space. Any vehicle configuration  $z \in Z$ is defined by $\{x(t), y(t),\theta(t), v(t)\}$, where $t$ denotes time in seconds, $x \in \mathbb{R}$  denotes the  $X$-coordinate in meters, $y \in \mathbb{R}$ denotes the $Y$-coordinate in meters, $\theta \in [-\pi,\pi] $ denotes the yaw in radians, and $v \in [0,v_{max}]$ denotes the absolute vehicle speed in meters per second.
The control input $u\in U$ is defined by $\{a(t), b(t),s(t), g(t)\}$, where $a \in [0,1]$ denotes the acceleration command, $b \in [0,1]$ denotes the brake command, $s \in [s_{min}, s_{max}]$ denotes the steering command in radians, and $g \in \{1,2,3\}$ denotes the gear command (reverse, neutral, forward) respectively.
The implemented RRT has a tree data structure denoted as $\mathbb{G}=(V,E)$, where $V$ or $z \in Z$ denotes the node and $E$ denotes the set of edges and is a subset of $Z$. The kinodynamic RRT implemented is described in {\bf Algorithm~\ref{algo:RRT}}.  

\begin{algorithm}
    \KwInput{$z_{init} = [x_{init}, y_{init}, \theta_{init},  v_{init}]$, \quad \quad $z_{goal} = [x_{goal}, y_{goal}, \theta_{goal}, v_{goal}]$, $Motion\_primitive\_set = \mathbb{S}$ = $[\{Z_1,U_1\},..\{Z_n,U_n\}]$ } 
   \KwOutput{Motion sequence or plan}
    
    $\mathbb{G} \gets \{z_{init} \}$\\
    \While{i=1 to N:}
   {
   $z_{rand}=Sample()$
   \newline
   \If {$p_{g}> random[0,1]$} 
   {
        $N_{near} \gets Nearest\_neighbour(\mathbb{G},z_{goal})$
    }
    \Else
    {
    $N_{near} \gets Nearest\_neighbour(\mathbb{G},Z_{rand})$
    }
    $z_{new}$,  $E_{new} \gets Extend (N_{near}, \mathbb{G}, \mathbb{S})$ 
    
    \If {Collision Checker is False} 
   {
   $\mathbb{G} \gets \{z_{new}$, $E_{new}\}$
    }
   \If {$Distance\_func(z_{new},z_{goal}) \leq Threshold$} 
   {
   Return $Tree\_Search(z_{new}, \mathbb{G})$
    }
   }
   \caption{Kinodynamic RRT with motion primitives}
   \label{algo:RRT}
    
\end{algorithm}

\paragraph{Sampling}
 RRT can be implemented with random sampling from uniform distribution. Areas near small openings in configuration space can be sampled more densely using Gaussian \cite{boor1999gaussian} distribution or Connectivity expansion \cite{kavraki1994randomized}. For simplicity we have implemented random sampling.
To control the expansion of tree, the random distribution range of $x$ and $y$ states can be manipulated. This is useful for maneuvers where a dense exploration on a small subspace of $Z$ is desired for applications such as  parallel parking, perpendicular parking, {\em etc.}

\paragraph{Goal Biasing} Goal Biasing ensures that the tree growth is directed towards the final goal so that a feasible path can be found within shorter time intervals. This is quite useful for real-time systems. Goal biasing can be tuned by a goal biasing probability $P_g$. In our implementation, $P_g$ is set to 0.2.

\paragraph{Distance function}
The distance function implemented is a weighted Euclidean distance, except for the $\theta$ dimension which is the geodesic distance in SO(2) as shown in \ref{Eq:dist_1} and \ref{Eq:dist_2} below.
\begin{equation} \label{Eq:dist_1}
   dist_{\theta_{so2}}  =min(abs(\theta_1-\theta_2), 2 \pi-abs(\theta_1-\theta_2)) 
   \end{equation}
 \begin{equation} \label{Eq:dist_2}
   dist=0.2 \sqrt{((x_1-x_2)^2+(y_1-y_2)^2)} + 0.8 \:dist_{\theta_{so2}}
 \end{equation}
 The distance function is required for the nearest neighbour search and also for terminating the algorithm if a feasible path within a threshold distance of goal is found. The threshold value used in this implementation is 0.2.

\paragraph{Nearest neighbour search}
The nearest neighbour search is critical for the time complexity of algorithm. There are many approximate and exact solutions to this problem. One of the efficient approaches is using a $K$-$d$ search \cite{cormen2009introduction}. In our approach, we implemented a naive brute force search. For the experiments in this study, the computation complexity was not an issue and the RRT planner was able to generate feasible trajectories within a small fraction of a second.

\paragraph{Extend}
The extend function is used to grow the tree towards the generated random configuration $z_{rand}$ from the nearest neighbour. In vanilla kinodynamic RRT \cite{lavalle2006planning, LaVelle2001}, the dynamics are solved with the given constraints to find feasible trajectories. We have used motion primitive to substitute the dynamic equations. Our implementation selects a random motion primitive from the available set and truncates it randomly along its length. Since motion primitives are defined in the vehicle frame of reference, the truncated motion primitive is transformed to the nearest node frame of reference. 

\paragraph{Collision checker}
The collision checker is required to confirm if the generated nodes and edges are obstacle free. If no obstacles are found along a node and edge pair, then it is appended to the tree. Nodes have a single configuration, so collision checker just needs to confirm whether that configuration is in $Z_{free}$. Edges have infinite number of configurations, so we have used an approximate method of collision check called dynamic checking \cite{lavalle2006planning}. This technique discretizes the edge at some fixed resolution $\epsilon$ and performs collision checks on those discrete configurations. 

We have implemented two types of collision checkers:  virtual collision checker and online mapped collision checker. The virtual collision checker creates simulated obstacles and performs collision checks with the created obstacles and the input configuration. It is implemented via python Shapely library \cite{gillies2013shapely}. This was required to create feasible plans for parallel parking and perpendicular parking. The online mapped collision checker uses the live map generated by the hector SLAM to generate feasible paths. The obstacle information in hector SLAM is stored as an occupancy grid (2D) of resolution $\delta= 0.05m/pixel$ along each dimension. Since the vehicle is not a point object, the query must be performed for all the pixels occupied by the vehicle, analogous to Minkowski sum \cite{hadwiger1950minkowskische}. To simplify this, the vehicle is assumed as a rectangle in the occupancy grid and 16 points on the outer edges are queried for any obstacle detection. This reduces the time complexity and enables generating feasible plans in real-time.

\paragraph{Tree search}The tree search returns a motion primitive sequence and  way-points from the goal configuration to the start configuration. The tree growth is terminated if the new node added is within 0.2m radius from the goal. The tree search simply traces the path from the nearest leaf node to the root node.

\subsection{Trajectory Tracking and Control}
We consider four basic actions to generate control sequence required for the vehicle to negotiate an obstacle-free trajectory, namely,
\begin{itemize}
    \item {\bf Vehicle gear control} which assumes values in {\it neutral, forward, reverse}
    \item {\bf Vehicle brake control} which is a switching control between ON and OFF and when set to OFF completely STOPS the vehicle to assume zero speed
    \item {\bf Steering angle control} with angular direction in {\it clockwise (-), counter-clockwise (+)} and with angular displacement at $10$ radians
     \item {\bf Velocity control} which sets the forward velocity within the range $(0, v_{max}]$
\end{itemize}

\begin{figure}[!ht]
    \centering
    \includegraphics[width = 7cm]{"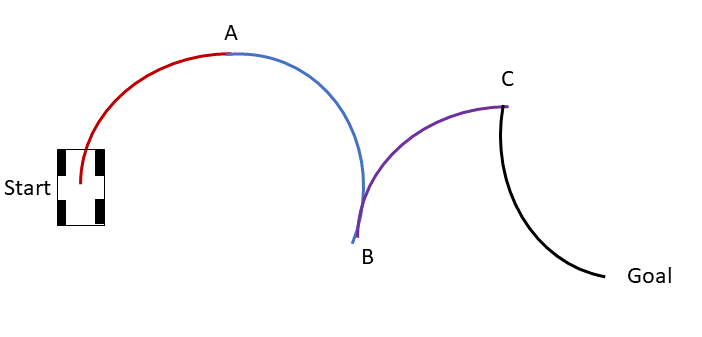"}
    \caption{Desired trajectory in terms of motion primitives }
    \label{trajectory1}
\end{figure}

Using the motion primitives, the RRT algorithm outputs a trajectory sequence from Start position to Goal while respecting the obstacle avoidance constraints. For example, consider an illustrative trajectory shown in \ref{trajectory1} suggested by the RRT algorithm. 
The suggested trajectory is composed of way-point segments connecting positions Start to A, A to B, B to C, and C to the goal in the configuration space. In terms of the motion primitives, the car is expected to follow the sequence 
\begin{equation} 
C_s = \{\text{{\it forward-right,forward-right,reverse-left,forward-left}} \} 
\end{equation}
to reach the goal through points A, B, and C. The actions in $C_s$ are executed as follows:
\begin{itemize}
    \item {\bf forward right from Start to A}:  Gear: forward, Steering: + 10 rad
    \item {\bf forward right from A to B}:  Gear: forward,  Steering: + 10 rad
    \item {\bf reverse left from B to C}: Gear: reverse, Steering: - 10 rad
    \item {\bf forward left from C to Goal}: Gear: forward, Steering: - 10 rad
\end{itemize}
With the values for Gear and Steer set as above, it remains only to control the speed of the vehicle.The PID controller controls the absolute variation in speed:
\begin{equation}
e_v(t) = |(v^*(t) -v(t))|
\end{equation}
where $v^*$ is the target speed and the $v$ is the actual speed of the vehicle.
As the car navigates, Hector SLAM provides the current position $p_x, p_y$  and the speed of the vehicle. The values are compared against the desired values for the $N$ way-points specified by $\{ x^*(t), y^*(t), \theta^*(t) , t \in \{t_1,t_2,..., t_N\}\}$ which will be used to derive error values at any time $t$ and drive the correcting control mechanism. In our experiment, we set the desired speed at a constant value (0.4m/s).
The PID controller with proportional, integral, and derivative gains, $K_p, K_I$, and $K_d$ respectively executes the following control:
\begin{equation}
u_{t_{n+1}} = K_p e_v(t_n) + K_I  \sum_{j=0}^{n} e_v(t_j) + K_d (\frac{e_v(t_n)-e_v(t_{n-1})}{\Delta t}) 
\end{equation}
where $t_n-t_0$ denotes the temporal interval buffer over which the error is integrated.
\section{EXPERIMENTAL RESULTS}
All experiments are performed in POLARIS GEM e2 autonomous vehicle platform shown in \ref{fig:vehicle}. It is equipped with Velodyne VLP-16 LiDAR used for localization and mapping, dual NVIDIA RTX 2080Ti GPU for GPU-accelerated edge computing, and drive by wire systems for steering, brake, gear, and acceleration controls. The entire communication interface is set up using ROS noetic \cite{quigley2009ros}. For simplicity, the initial configuration is always (0,0,0,0) and the goal configuration is provided as per the experiment. Table \ref{tab:plan} compares the desired goal configuration and the final configuration generated by the planner. The error or drift from the final configuration is computed using the distance metric presented in Equations \ref{Eq:dist_1} and \ref{Eq:dist_2}. Table \ref{tab:drift} summarizes the drift values observed for all the experiments. For the videos of different navigation maneuvers, click  \href{https://drive.google.com/drive/folders/1NzYhRjUM0KSE0MDjicWkTqcjokVSojSz?usp=sharing}{HERE}
\begin{table}[H]
\caption{Plans generated}
\label{tab:plan}
\centering
\begin{tabular}{c||c||c||c||c}
\hline
\bfseries \begin{tabular}{@{}c@{}}Experiment\end{tabular} & \begin{tabular}{@{}c@{}}Planning input\\ and output\end{tabular} &\bfseries \begin{tabular}{@{}c@{}}X \end{tabular} & \bfseries \begin{tabular}{@{}c@{}}Y\end{tabular} & \bfseries \begin{tabular}{@{}c@{}}$\Theta$ \end{tabular} 
\\
\hline\hline
 {Reverse Navigation} & Goal & -11 & 2 & $\pi/4$ \\
& Output & -10.6 & 1.6 & 0.64 \\
\hline
{Parallel parking trial 1} & Goal & 4 & 3 & 0 \\
& Output & 4.4 & 3.4 & -0.02 \\
\hline
 {Parallel parking trial 2} & Goal & 4 & 2  & 0 \\
& Output & 3.8 & 1.9 & -0.02 \\
\hline
 {Perpendicular parking} & Goal & 4 & 3 & $\pi/2$ \\
& Output & 4.1 & 3.3 & 1.54 \\
\hline
\end{tabular}
\end{table}

\begin{table}[H]
\caption{Error or drift from final configuration}
\label{tab:drift}
\centering
\begin{tabular}{c || c } 
\hline
\bfseries \begin{tabular}{@{}c@{}}Experiment\end{tabular} &  \begin{tabular}{@{}c@{}}error in m \end{tabular} 
\\
\hline\hline
 {Online Navigation} & 0.05 \\
\hline
 {Parallel parking trial 1} & 0.6 \\
\hline
 {Parallel parking trial 2} & 0.5 \\
\hline
 {Perpendicular parking} & 0.2 \\
\hline
\end{tabular}
\end{table}
\subsection{Reverse Navigation in a Constrained Environment}
  
\begin{figure}[!ht]
\centering
   \includegraphics[width=1\linewidth]{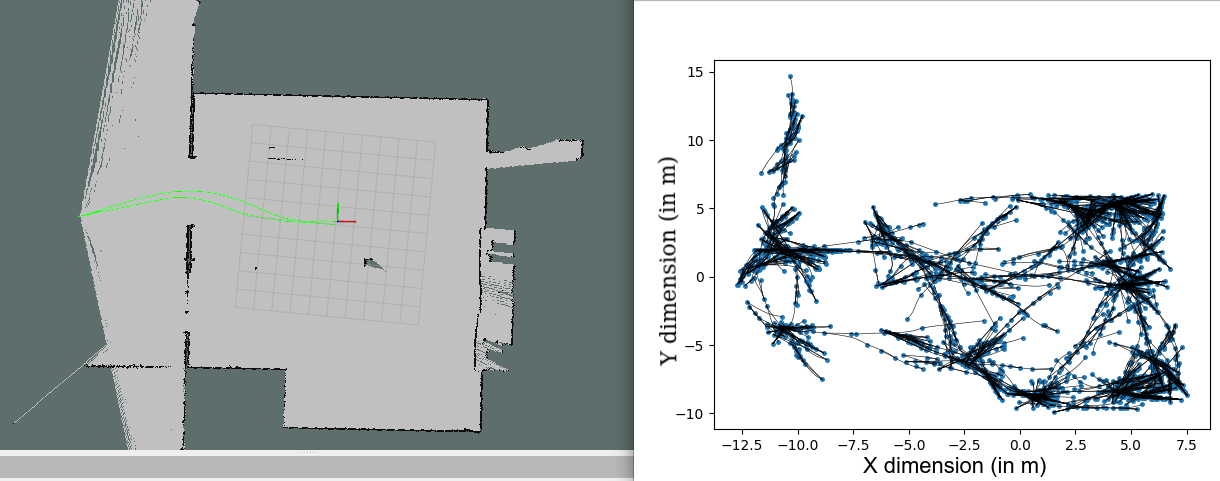}
   \caption{Reverse navigation in a constrained environment}
   \label{fig:Live_map_1} 
  \end{figure}

For this experiment, the objective was to perform path planing with live obstacles generated from LiDAR map. Since LiDAR cannot detect small obstacles, this experiment was performed inside the building (high-bay) with goal to reverse starting from the original inside location (taped boundary) without crashing into any obstacles or wall boundaries. Figure \ref{fig:Live_map_1} shows the high-bay LiDAR mapping and the RRT map generated for this experiment. The final desired orientation was $\pi/4$ radians. This path consisted of 7 way-points connected by reverse primitives. In terms of SLAM accuracy, since the experiment was conducted indoors, the LiDAR was able to get good number of returns enabling accurate localization and tracking. Hence, in terms of drift, the vehicle followed the path with very little/no drift (Table \ref{tab:drift}) . 

\subsection{Parallel Parking}
For this experiment, the goal was to successfully execute parallel parking starting from a street view orientation (starting orientation is the same angle as the final parked orientation) following a path such that it avoids any collisions with cars parked in front and behind the goal parking location. Figure \ref{fig:parallel_plan_1} shows the RRT graph and the trajectory generated towards the goal. The RRT graph shows all feasible trajectories that avoid the two simulated obstacles. Also, Figure \ref{fig:parallel_plan_1} and Figure \ref{fig:parallel_plan_2} show the random behaviour of the planner. It may be noted that even for the same obstacles in C-SPACE, the expansion of tree is not unique. Trial 1 shows a path consisting of a combination of forward and reverse primitives connecting 11 way-points. Since RRT is a random plan generator, we were able find a better path with fewer way-points by running the planner again (Trial 2). On the autonomous vehicle platform, we successfully demonstrated the feasibility and tracking for both the plans generated by RRT. Since these plans were generated by the motion primitive set, the tracking was possible through  predefined control inputs and the linear PID controller for disturbance rejection.

\begin{figure}[ht!]
\centering
\begin{subfigure}[b]{0.55\textwidth}
 \centering
   \includegraphics[width=0.65\linewidth]{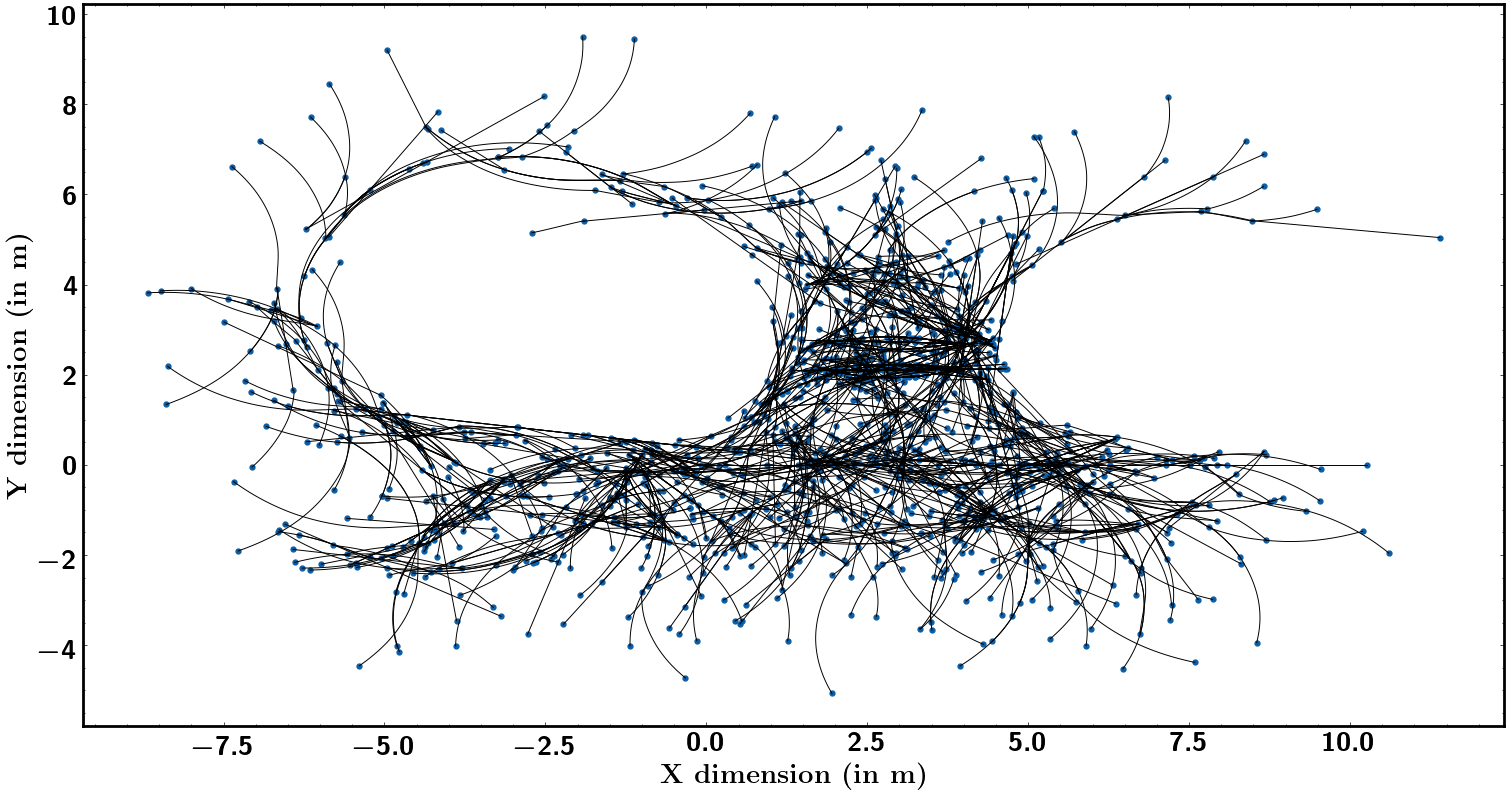}
   \caption{}
   \label{fig:parallel_plan_1} 
\end{subfigure}
\begin{subfigure}[b]{0.55\textwidth}
\centering
   \includegraphics[width=0.65\linewidth]{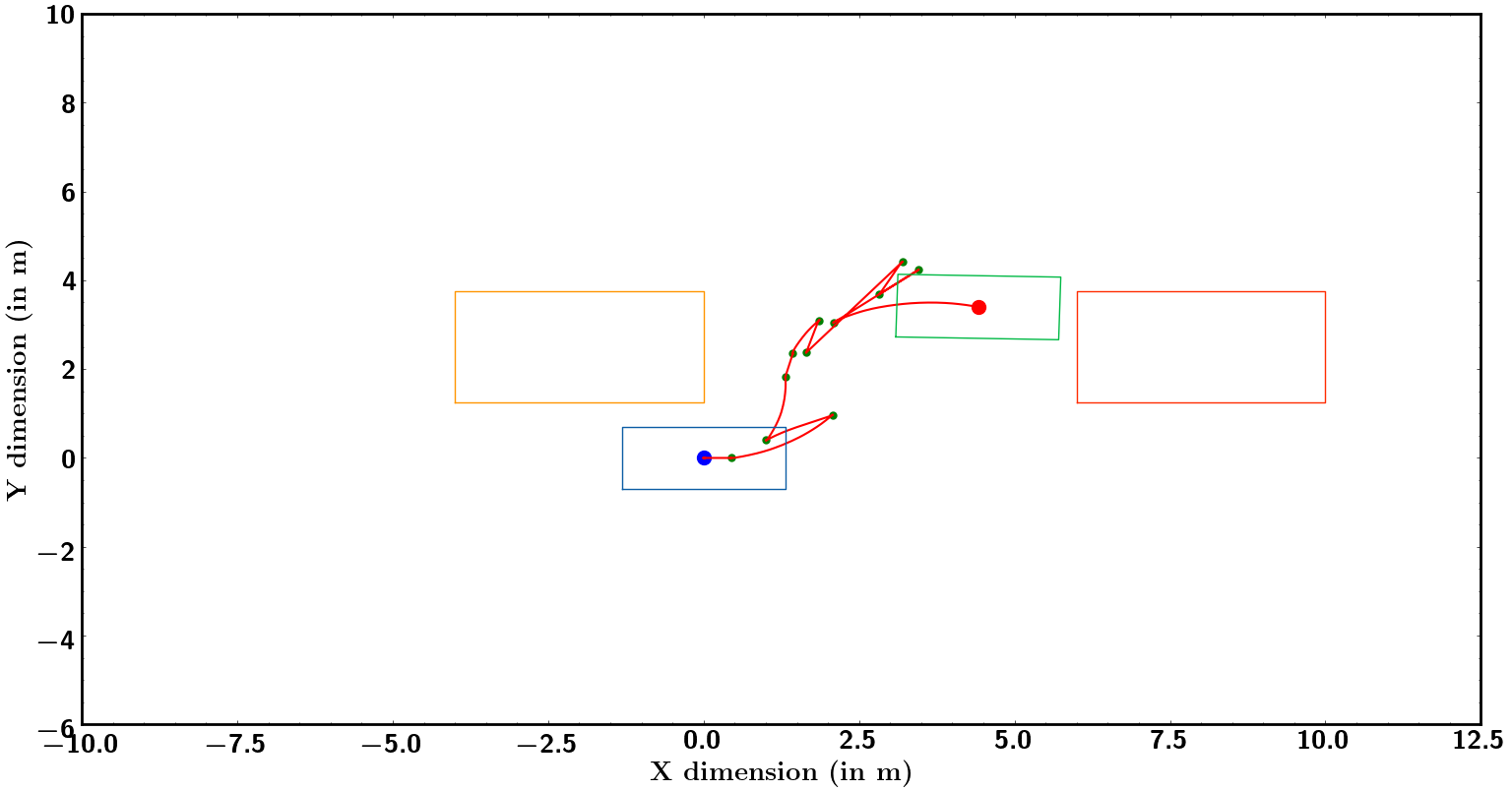}
   \caption{}
   \label{fig:Ng2}
\end{subfigure}
\caption{Parallel Parking Trial 1 (a) RRT graph (b) Generated Trajectory}
  \label{fig:parallel_plan_1}
\end{figure}
\begin{figure}[H]
\centering
\begin{subfigure}[b]{0.55\textwidth}
 \centering
   \includegraphics[width=0.65\linewidth]{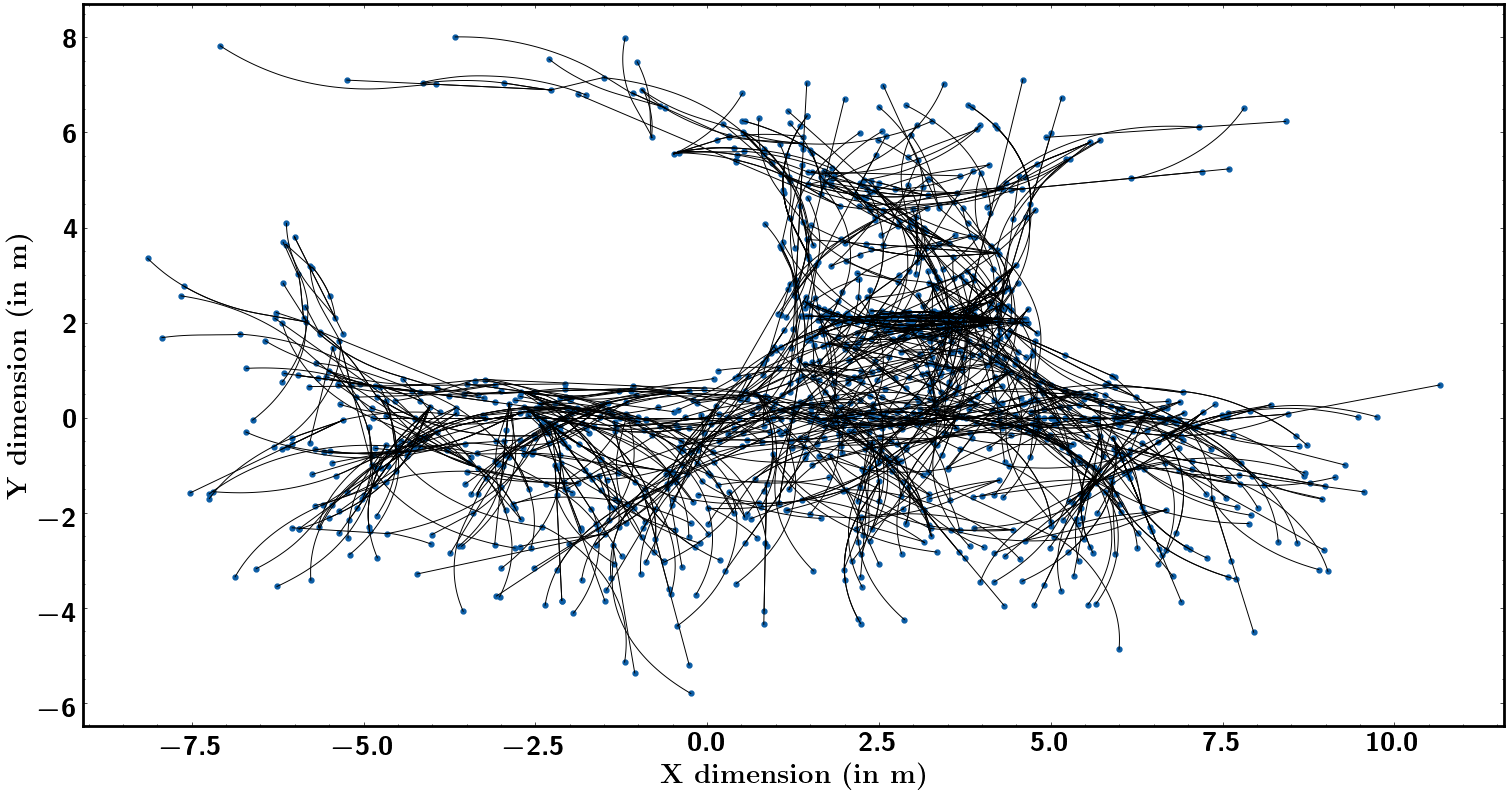}
   \caption{}
   \label{fig:parallel_plan_2} 
\end{subfigure}
\begin{subfigure}[b]{0.55\textwidth}
\centering
   \includegraphics[width=0.65\linewidth]{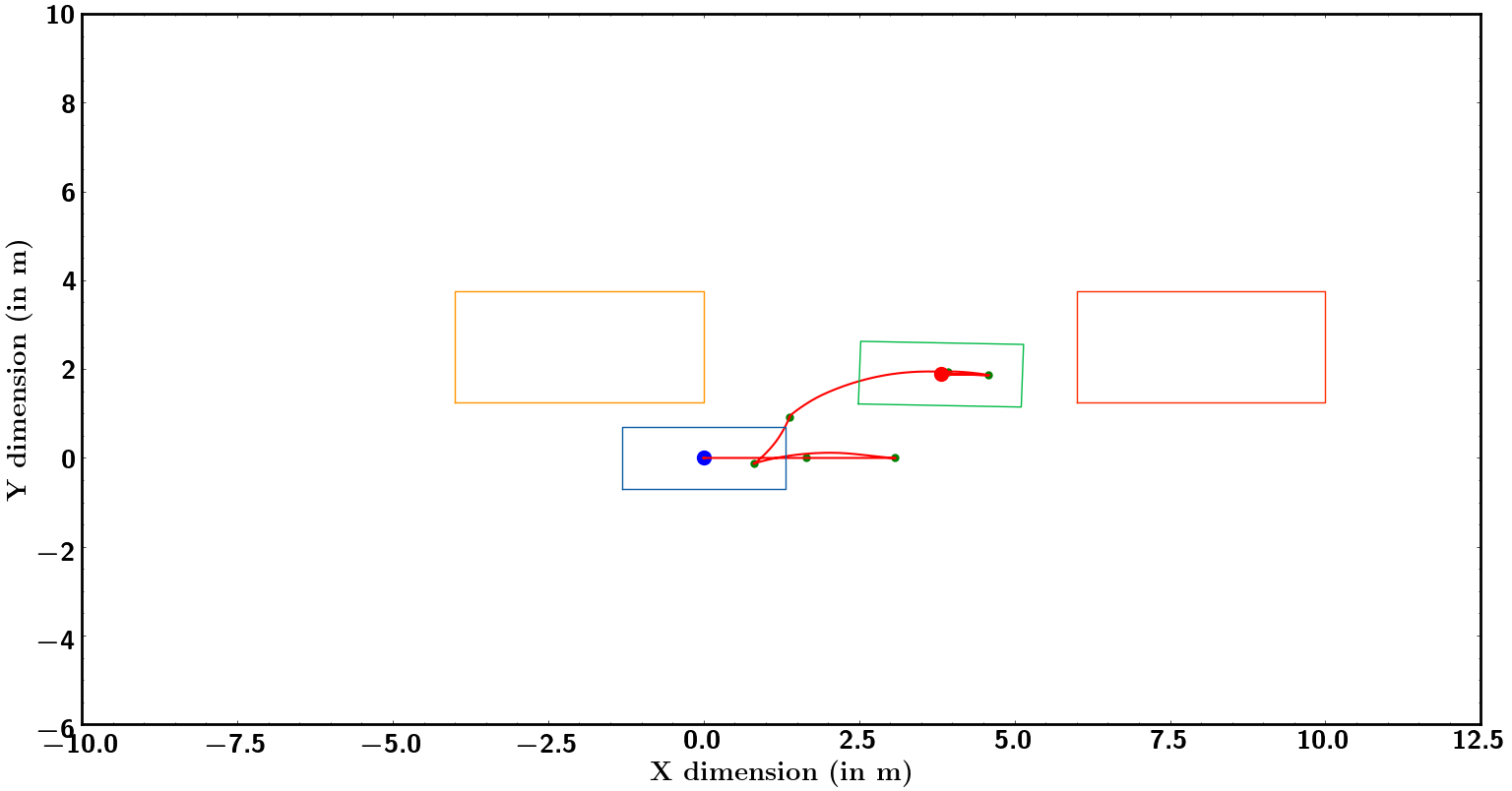}
   \caption{}
   \label{fig:Ng2}
\end{subfigure}
\caption{Parallel Parking Trial 2 (a) RRT graph (b) Generated Trajectory}
  \label{fig:parallel_plan_2}
\end{figure}
 In terms of SLAM accuracy and drift, this experiment ended up with some deviation from the desired goal (Table \ref{tab:drift}). This could be probably due to the open space on one side of the vehicle which does not provide enough LiDAR returns for accurate SLAM estimations.

\subsection{Perpendicular Parking}
For this experiment, the goal was to successfully park perpendicularly in a parking spot starting from a street view orientation (starting orientation of the car is a 90 degree rotation from final parked orientation) following a path that avoids any collisions with cars parked to the left and right of the goal parking location. From Figure \ref{fig:perp_park}, we can observe that the vehicle followed a path consisting of 6 way points. The motion consisted mostly of the forward motion, other than the reverse motion from way-point 3 to way-point 4 to fix the orientation and avoid collision with a possible vehicle to the right of the goal parking spot. In terms of tracking, there was slight drift due to SLAM inaccuracies, similar to parallel parking.
\begin{figure}[ht!]
\centering
\begin{subfigure}[b]{0.55\textwidth}
 \centering
   \includegraphics[width=0.65\linewidth]{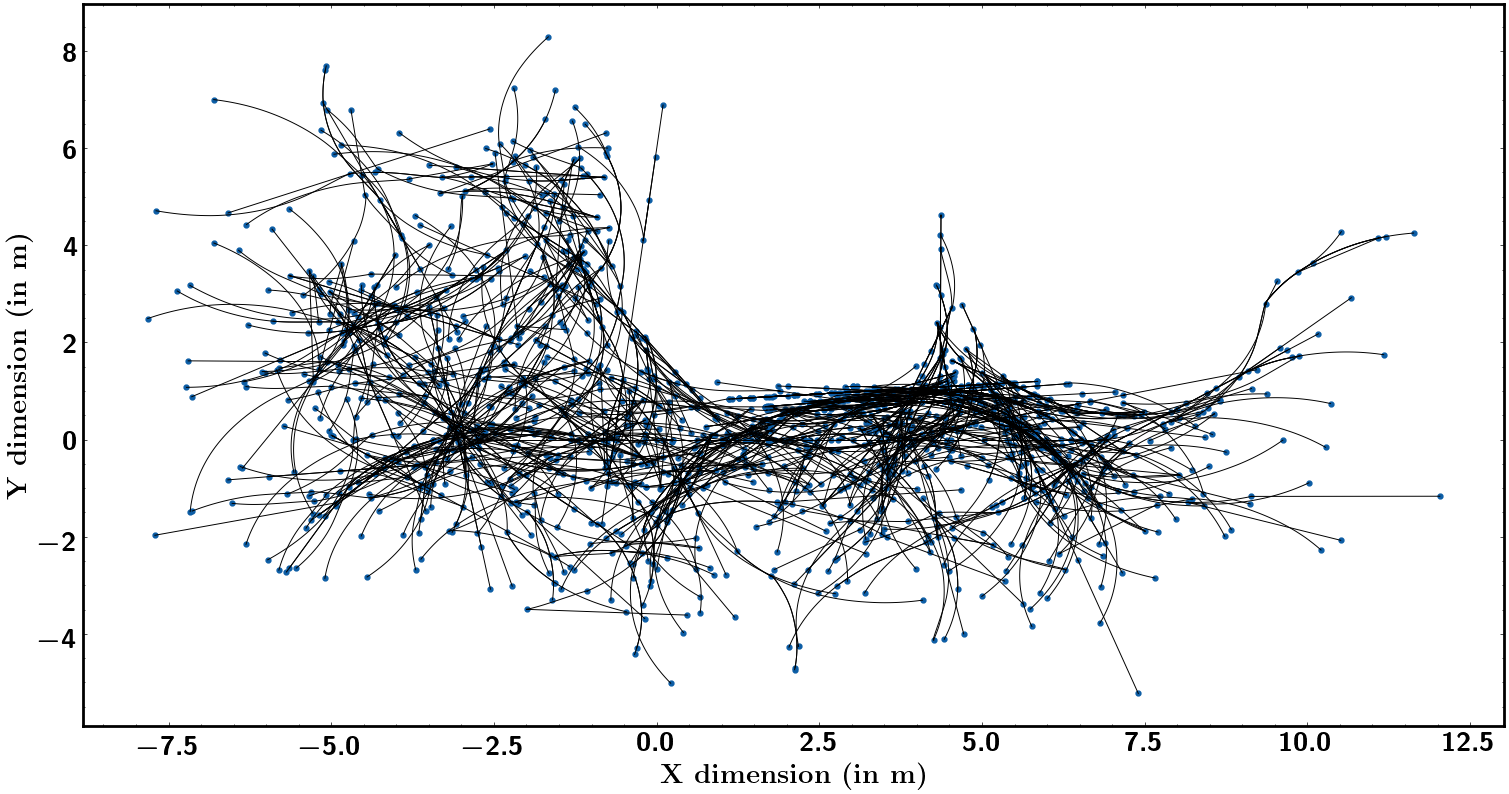}
   \caption{}
   \label{fig:perp_park} 
\end{subfigure}
\begin{subfigure}[b]{0.55\textwidth}
\centering
   \includegraphics[width=0.65\linewidth]{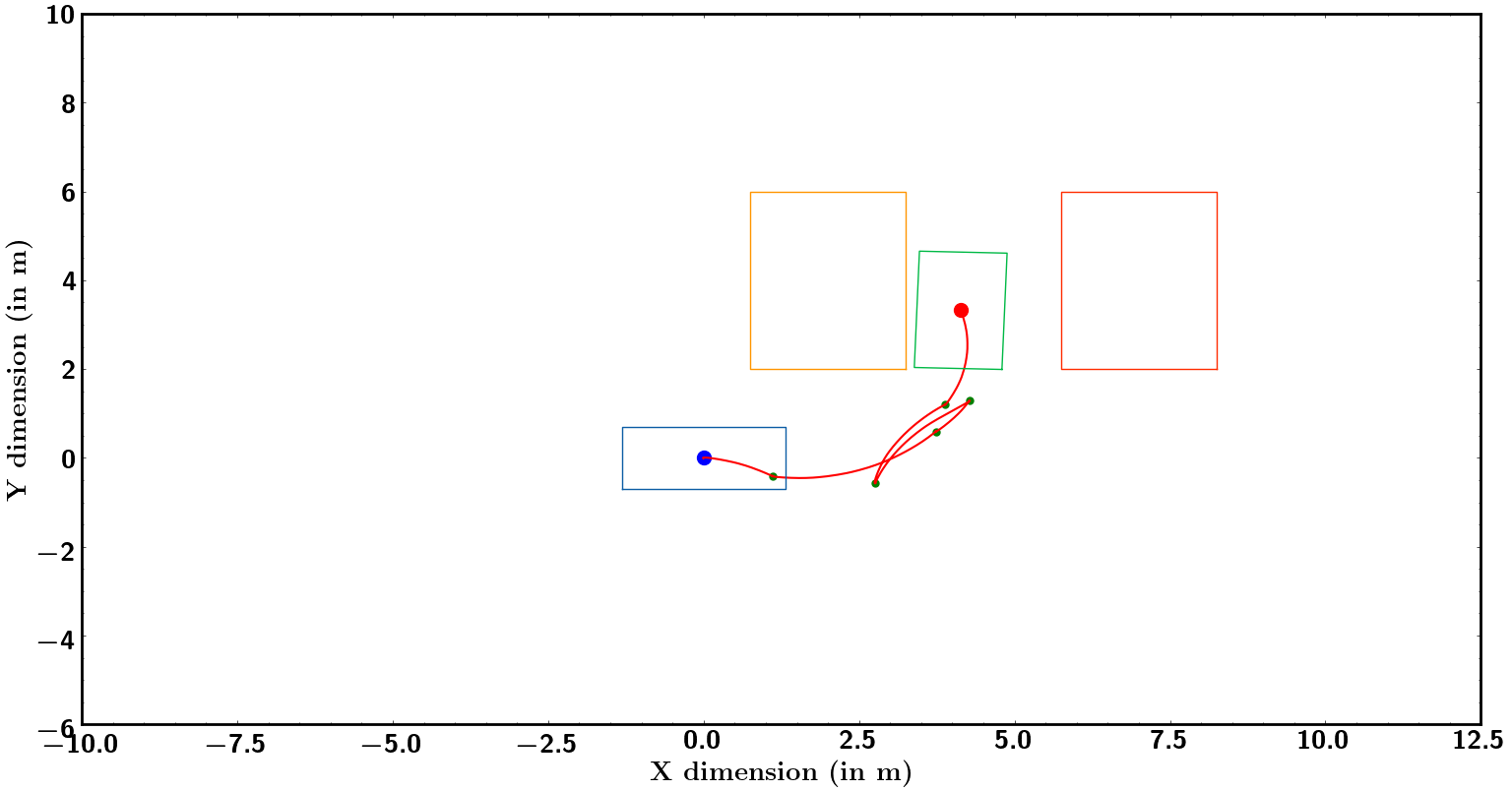}
   \caption{}
   \label{fig:perp_park}
\end{subfigure}
\caption{Perpendicular parking (a) RRT graph (b) Generated Trajectory}
  \label{fig:perp_park}
\end{figure}

\section{CONCLUSIONS}
In the present work, we explored a motion primitive based kinodynamic RRT planner to design obstacle free trajectories in C-SPACE for non-holonomic systems such as autonomous vehicles. For localization and mapping, we used Hector SLAM and implemented tracking and control using linear PID and SLAM feedback. The methodology successfully led the vehicle to perform complex maneuvers like parallel parking, perpendicular parking, and online navigation. We observed that for indoor navigation, tracking errors were negligible but for outdoor experiments like parallel and perpendicular parking the tracking errors compounded along the trajectory. We wish to apply sensor fusion with an IMU  to control tracking errors. The study can be extended to dynamic unstructured environments and different robotic systems by learning the motion primitives directly using machine learning techniques and continual adaptation. Also, the planner can be upgraded to find solutions using cost-based expansion approaches similar to $RRT^*$\cite{karaman2011sampling}.

\addtolength{\textheight}{0cm}   




\section*{ACKNOWLEDGMENT}

We sincerely thank Prof David Forsyth for giving us access to the experimental environment, and our colleagues Ruhi, Didrick, and Rahul for their support in carrying out the navigation experiments.

\bibliographystyle{ieeetr}
\bibliography{refs.bib}

\end{document}